# Artificial intelligence and pediatrics: A synthetic mini review

Peter Kokol[1]*, Jernej Završnik[2] and Helena Blažun Vošner[3]

[1]University of Maribor, Faculty of Electrical Engineering and Computer Science, Slovenia

[2]Dr. Adolf Drolc Healthcare Centre, Slovenia

[3]University of Maribor, Faculty of Health Sciences, Center for International Cooperation, Slovenia

## Introduction

The use of artificial intelligence intelligence (AI) in medicine can be traced back to 1968 when Paycha published his paper Le diagnostic a l'aide d'intelligences artificielles: presentation de la premiere machine diagnostri [1]. Few years later Shortliffe et al. [2] presented an expert system named Mycin which was able to identify bacteria causing severe blood infections and to recommend antibiotics. Despite the fact that Mycin outperformed members of the Stanford medical school in the reliability of diagnosis it was never used in practice due to a legal issue - who do you sue if it gives a wrong diagnosis? [3]. However only in 2016 when the artificial intelligence software built into the IBM Watson AI platform correctly diagnosed and proposed an effective treatment for a 60-year-old woman's rare form of leukemia the AI use in medicine become really popular [4].

On of first papers presenting the use of AI in paediatrics was published in 1984. The paper introduced a computer-assisted medical decision making system called SHELP, aimed to diagnose inborn errors of metabolism even in very rare cases [5]. More than 30 years later the above mentioned Watson platform was successfully utilized in Boston Children's Hospital to help their clinicians diagnose and treat rare paediatric diseases [6].

AI based decision support is founded on machine learning (ML), a subfield of computer science, defined by Arthur Samuel in 1959, as the ability to teach computers what to do without being explicitly programmed [7]. The advantage of machine learning is that it "learns" from data. Such data is routinely collected in electronic medical encounters. ML works well either on big data, however it is also more sensitive than traditional statistical methods on small data. Another advantage of ML is that it searches the data for all possible hypothesis, hence no a – priory hypothesis is needed [8].

## AI in pediatrics from a bibliometric perspective

To get a broader picture of the utilization of AI in paediatrics from the descriptive, historical and thematic point of view, we analysed the research literature retrieved from the Web of Science database (Thomas Reuters, USA) using a bibliometric approach. "Bibliometrics is to scientific papers as epidemiology is to patients" argued Lewisson and Devey in their 1999 paper [9]. From the scientific point of view bibliometrics could be defined as the application of mathematical, statistical and heuristic methods to scientific publications [10]. To analyse and visualize the context of the application of AI and ML in paediatrics we used bibliometric mapping and a tool called VOSviewer (Leiden University, Netherlands) [11]. A bibliometric map could reveal different patterns and aspects of research literature content in a form of science landscapes. In our study two types of landscape were induced, namely a timeline landscape and a cluster landscape. The timeline science landscapes was used to analyse the evolution of research topics. The terms appearing in the cluster landscape were used as codes to perform thematic analysis [12]. The codes were used first to define and name a theme for each cluster and second as keywords to select the appropriate papers from the corpus to refine themes.

The search in Web of Science All databases (Thomas Reuters, USA) performed on 26th of August resulted in a corpus of 1662 papers. Most of them were original papers and reviews (87%). The search was done using the search string *"machine learning" or "rough sets" or ("decision tree*" and (induction or heuristic)) or "artificial neural networks" or "support vector machines" or "rough sets" or "deep learning" or "intelligent systems" or "artificial intelligence"* restricted to the research area *Pediatrics*.

The analysis of the corpus showed that first paper presenting the application of AI in pediatrics was published in 1984 (Figure 1). Two years later a first article on use of machine learning appeared, introducing a classifier of ECG signals based on linear prediction techniques [13]. Incubation period in literature production, featuring a small and constant number of publications lasted till 1993 (n < 4 publication/year), when the initiation period begun. The initiation period was characterized by a linear growth in the number of publications (from 20 to 44 publication/year). In 2005 the rapid growth phase began (from 60 to 180 publications/year) followed by a decline in the number of publications after 2015. In 1993 when the positive trend started, the research literature production was spread between seven countries. In 2015, in the time of literature production peak, the production country spread reached 45 countries.

The timeline science landscape (Figure 2) shows three historical periods:

- **Before 2008**: Research was focused on applications based on decision trees, genetic algorithms, artificial neural networks and ruled based systems. They were used for knowledge extraction and decision making. The areas were preterm birth, mortality and

---

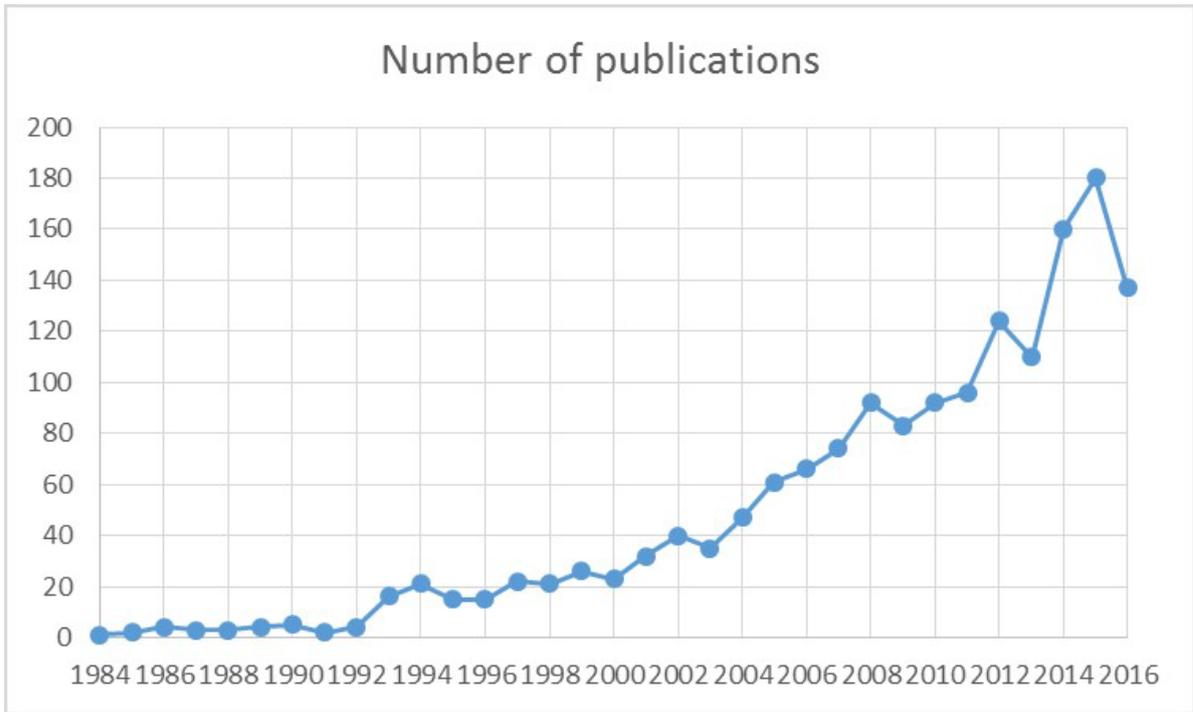

**Figure 1.** The dynamics of research literature production in artificial intelligence use in paediatrics

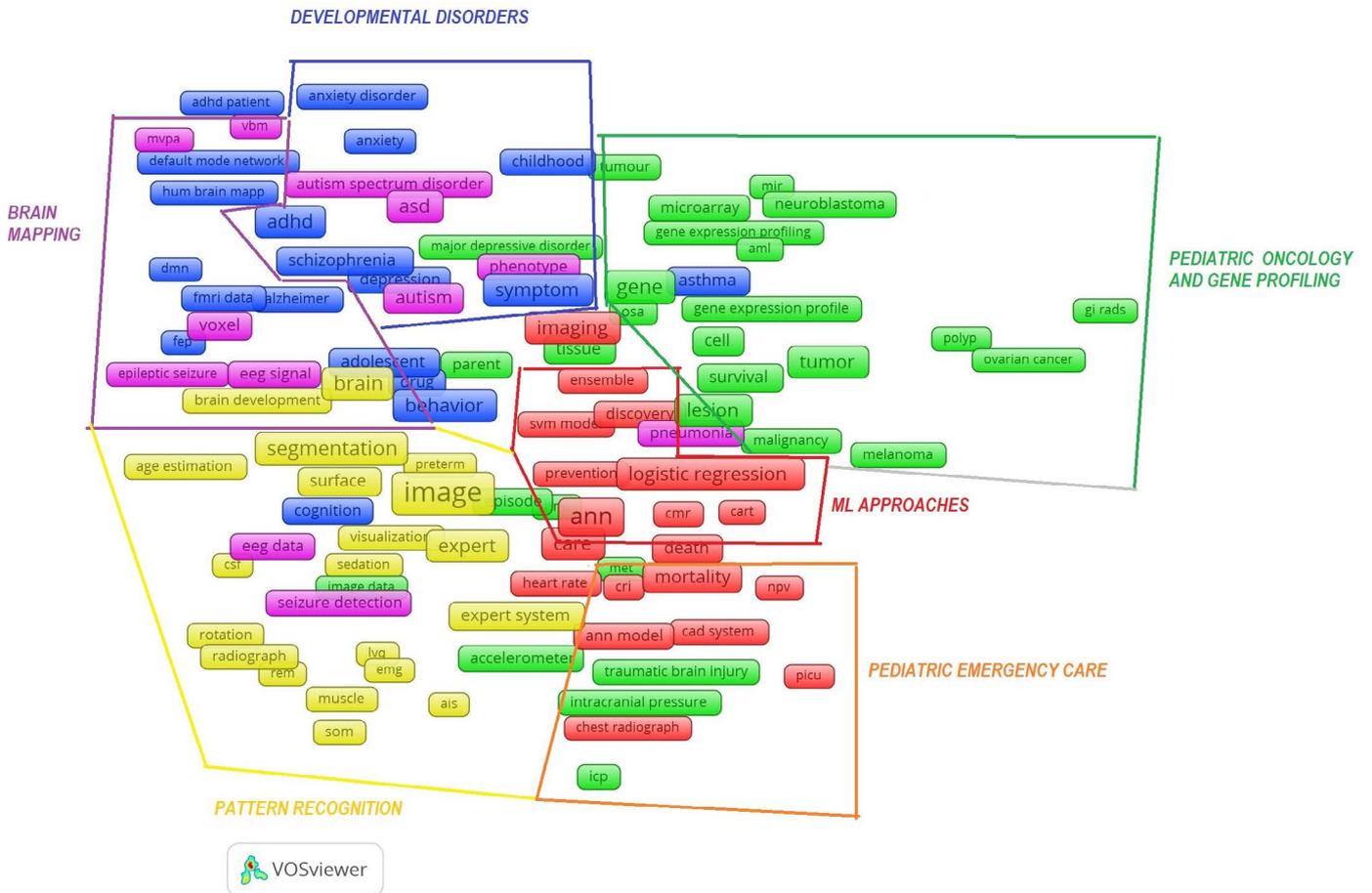

**Figure 2.** The timeline of research literature production on artificial intelligence use in paediatrics

survival prediction, and cancer, neuroblastoma, melanoma and lesion treatment;

- **2009 – 2012:** Application included use of discriminant analysis, logistic regression models and support vector machine for prediction, prognosis, therapy, care, feature selection, and signal (EEG, ECG) speech and image processing. Health areas concerned were infections, seizure, genetics and pathology. Target groups were new-borns, premature infants and young children. The robots and computer aided design system were introduced into paediatrics.

- **2013 and beyond:** This period was focused on application of ML in schizophrenia, pneumonia, asthma, abnormality and epilepsy. The period is also characterized with a new target group - the children with autism spectrum disorder and children with attention deficit hyperactivity. No new ML approaches emerged, however, the assessment of accuracy with introduction of new metrics become important. The research focus was shifted from classification to predictive models.

Using the thematic analysis (Figure 3) we identified six themes:

**AI in brain mapping** applications include prediction of child brain maturity based on fMRI [14], brain functional connectivity in preterm infants [15], classifying individuals at high-risk for psychosis based on functional brain activity [16], prediction of pediatric unipolar depression [17], analysis of resting-state brain function for attention-deficit/hyperactivity disorder, predicting the language outcomes following cochlear implantation [18] and similar;

**AI use in pattern recognition** are used for seizure prediction in children with epilepsy [19], visualization of complex data [20], predicting neurodevelopment [21] , identifying motor abnormalities [22], analysing EMG, ECG and other signal [23], image analysis, segmentation [24], etc;

**AI use in developmental disorders** where examples include quantifying risk for anxiety disorders in preschool children [25], developing socially intelligent robots as possible educational or therapeutic toys for children with autism [26], identifying children with autism based on face abnormality [27] etc;

**AI in paediatric emergency care** was used for automated appendicitis risk stratification [28], supporting diagnostic decisions [29], traumatic brain injury [30] and detection of low-volume blood loss [31];

**Machine learning approaches** most frequently employed were artificial neural networks [19], support vector machines [32], decision trees [33] and Bayesian networks [34]; and

**AI use in pediatric oncology and gene profiling was** characterized with applications like identification of regenerating bone marrow cell populations [35], benchmarking key genes for cancer drug development (36), gene expression profiling for children with neuroblastoma or lymphoblastic leukaemia [37].

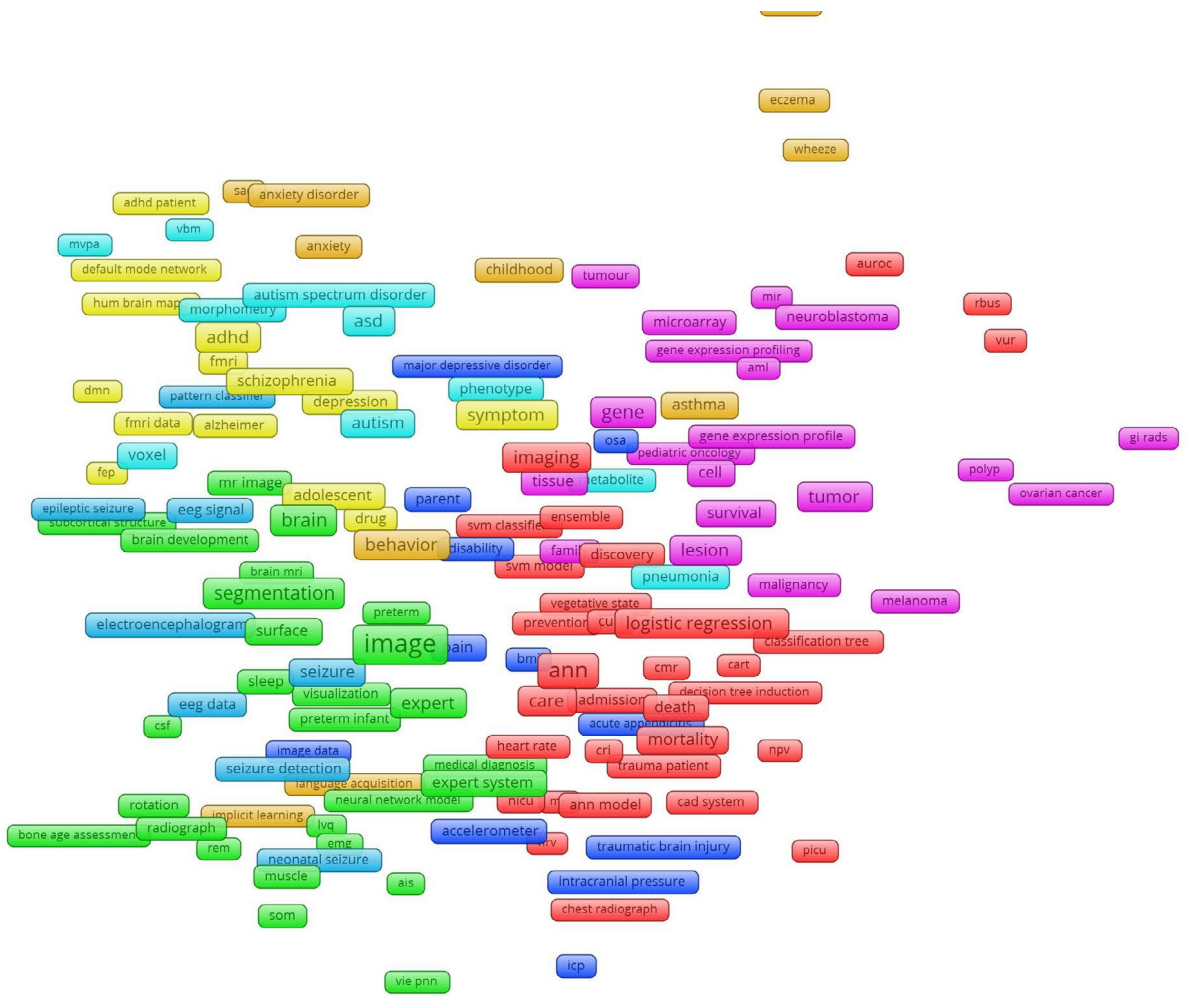

**Figure 3.** Themes regarding artificial intelligence use in paediatrics

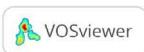

## Conclusion

Above analysis showed that AI use in pediatrics is expanding. Their use resulted in more accurate and faster diagnoses, improved decision making, more specific and sensitive identification high-risk and improved clinical outcomes. In that manner there were less adverse events, less readmissions and in general less costs. However the analysis also revealed that AI is still not used to their full potential. Main reasons encountered were:

- AI tools are not pediatrician friendly and requires specific information technology skills to be employed

- most pediatricians either have never even heard of AI, or they don't trust it;

- AI use has been mostly tested only on small cases; and

- there are legal issues in AI use in medicine.

To overcome above obstacles AI approaches must be made more user friendly. Academic institutions should support and reward the development of AI capability and capacity in medical students. Successful applications of AI in practice or research should gain academic and professional recognition and promotion. Academic pediatric centers should include AI use in their strategies for building better health care.

## Author contributions

Study concept and design: PK, JZ, HB. Interpretation of data: PK, JZ; HB. Drafting of the manuscript: PK. Critical revision of the manuscript: JZ, HB, Data analysis and acquisition of data: PK.